\documentclass[11pt,a4paper]{article}
\usepackage[table]{xcolor}
\usepackage[hyperref]{acl}

\usepackage{times}
\usepackage{latexsym}
\usepackage{amssymb}
\usepackage{amsmath}
\usepackage{graphicx}
\usepackage{float}
\usepackage{enumitem}
\usepackage{booktabs}
\usepackage{siunitx}
\usepackage[capitalise]{cleveref}

\usepackage{microtype}
\newcommand*{\affmark}[1][*]{\textsuperscript{#1}}
\newcommand*{\affaddr}[1]{#1}

\DeclareMathOperator*{\argmin}{arg\,min}
\usepackage{todonotes}

\newcommand{\mytt}[1]{{\tt{#1}}}

\makeatletter
\def\thanks#1{\protected@xdef\@thanks{\@thanks
        \protect\footnotetext{#1}}}
\makeatother

\title{When More Data Hurts: A Troubling Quirk in Developing Broad-Coverage Natural Language Understanding Systems}

\author{Elias Stengel-Eskin\affmark[1]\footnotemark[1]\thanks{\llap{\textsuperscript{*}}{Work done as an intern at Microsoft Semantic Machines.}} 
        \quad 
        Emmanouil Antonios Platanios\affmark[2]
        \quad 
        Adam Pauls\affmark[2]
        \AND
        Sam Thomson\affmark[2]
        \quad
        Hao Fang\affmark[2]
        \quad 
        Benjamin Van Durme\affmark[2]
        \quad 
        Jason Eisner\affmark[2] 
        \quad 
        Yu Su\affmark[2] \\ \\
        \affaddr{\affmark[1]Johns Hopkins University} \quad 
        \affaddr{\affmark[2]Microsoft Semantic Machines} 
        }    

\date{}

\begin{document}
\maketitle

\begin{abstract}
In natural language understanding (NLU) production systems, users' evolving needs necessitate the addition of new features over time, indexed by new symbols added to the meaning representation space. This requires additional training data and results in ever-growing datasets.
We present the first systematic investigation of this \textit{incremental symbol learning} scenario. 
Our analysis reveals a troubling quirk in building broad-coverage NLU systems: 
\textit{as the training dataset grows, performance on the new symbol often decreases if we do not accordingly increase its training data.} 
This suggests that it becomes more difficult to learn new symbols with a larger training dataset.
We show that this trend holds for multiple mainstream models on two common NLU tasks: intent recognition and semantic parsing.
Rejecting class imbalance as the sole culprit, we reveal that the trend is closely associated with an effect we call \textit{source signal dilution}, where strong lexical cues for the new symbol become diluted as the training dataset grows.
Selectively dropping training examples to prevent dilution often reverses the trend, showing the over-reliance of mainstream neural NLU models on simple lexical cues.\footnote{Code, models, and data are available at \url{https://aka.ms/nlu-incremental-symbol-learning}.}

\end{abstract}

\section{Introduction} 
\label{sec:intro}
Broad-coverage natural language understanding (NLU) systems that simultaneously support a wide range of user requests are critical for developing general-purpose natural language interfaces. 
Such systems are already being deployed and reach millions of users worldwide: As of 2021, Amazon Alexa contains more than \num{80000} skills \citep{vailshery_2021}, and Microsoft has deployed a new conversational interface for Outlook that uses over \num{300} composable functions to represent fine-grained semantics in task-oriented user-agent dialogues \cite{andreas.j.2020, outlook_2021}.

These broad-coverage NLU systems do not acquire their full capability on day one: new features (e.g., intents or functions), and the new training examples for learning them, are added incrementally. However, there has been little research on the data and learning dynamics during such incremental development. The wide deployment and high cost of such systems make this research critical, and we aim to call attention to this important problem. 
\begin{figure}[t]
    \centering
    \includegraphics[width=\linewidth]{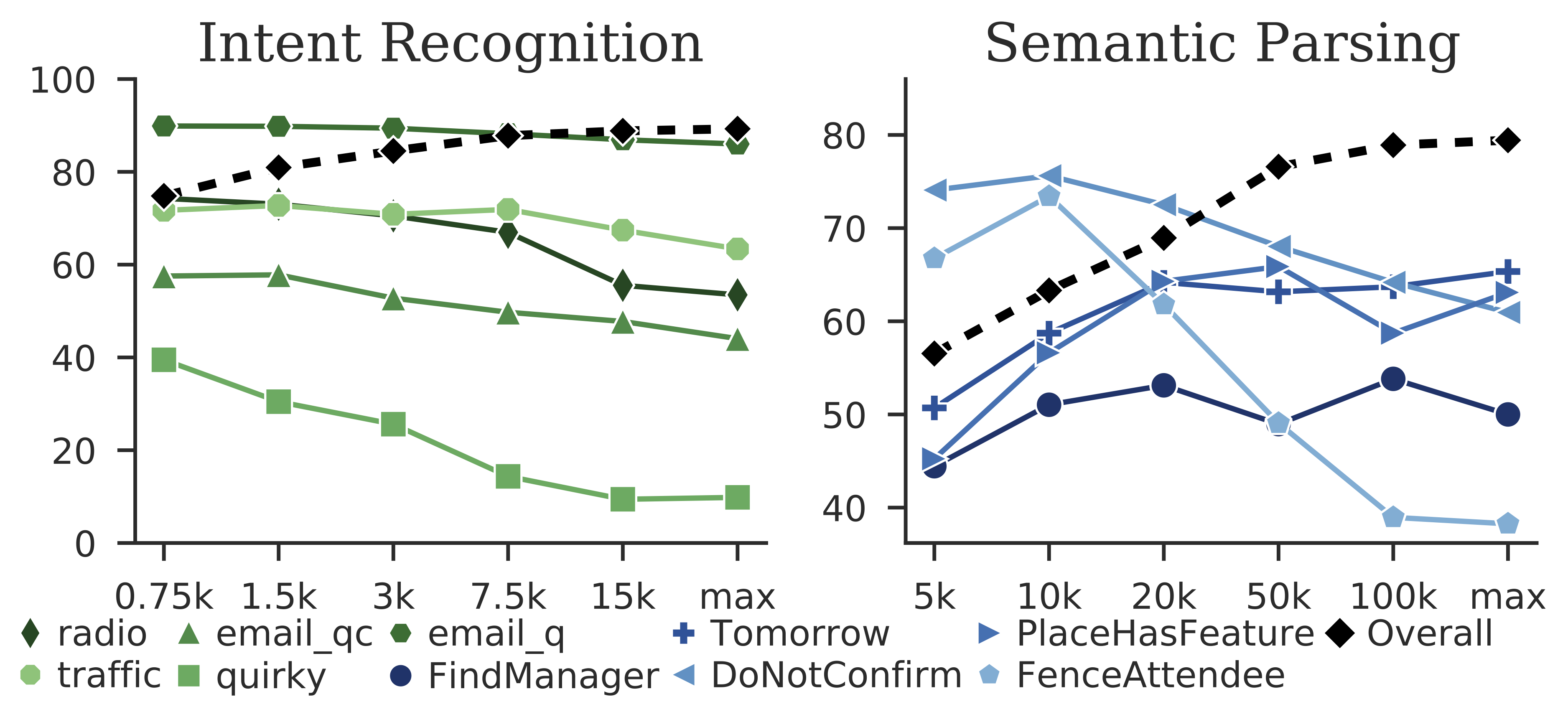}
    \vspace{-2em}
    \caption{Overall and per-symbol test accuracy for learning a new symbol from a fixed number of training examples (30 for intent, 100 for SMCalFlow) when the overall training dataset grows (x-axis). Each line represents fixing a different symbol. As training data size increases, overall accuracy (averaged across symbols) increases but new symbol accuracy often decreases.} 
    \label{fig:intent_and_calflow}
    \vspace{-2em} 
\end{figure}

We consider two prototypical NLU tasks: intent recognition and semantic parsing. 
NLU is generally concerned with mapping utterances into symbolic meaning representations (e.g., intent labels for intent recognition and sequences of functions/arguments for semantic parsing). 
We consider the following \textit{incremental symbol learning} scenario: given a set of existing symbols and their training data, we want to learn a new symbol, which entails adding new labeled data for the symbol. 
As the system supports more and more symbols, its training data size continually increases. 
This relates more broadly to domain adaptation \citep{ben-david.s.2010} if we consider the new symbols to be a new domain with a very small number of examples, and differs from continual learning~\cite{madotto-etal-2021-continual}, which learns a new domain using only training data for the new domain (with replay of limited data from previous domains). 
We instead train models from scratch on all the existing data and the new symbol data simultaneously, which is typical in practice to guarantee accuracy on both old and new features. 

At first blush, dataset growth may seem positive, in holding with a common assumption of supervised learning: \textit{more data is generally better} \citep{kearns.m.1994}. 
However, our analyses reveal a troubling quirk: \textit{as the training data size increases, the performance on the new symbol decreases}. 
To investigate this, we simulate the incremental symbol learning scenario by trying to learn a new symbol from a fixed number of new training examples when the overall training data size gradually increases. 
We examine an English intent recognition dataset \citep{liu.x.2019} and an English semantic parsing dataset \citep{andreas.j.2020},  testing 5 symbols per dataset.

Fig.~\ref{fig:intent_and_calflow} shows the overall test accuracy of our best models (cf.\ \S\ref{sec:data_and_model}) as well as the accuracy on examples containing the new symbol.
As the dataset grows, the average test accuracy across all symbols increases monotonically. 
However, the accuracy on the new symbol's examples generally decreases. 
This decrease could lead to an increased demand for annotations, with more training examples needing to be collected to achieve adequacy for each new symbol as the dataset grows. 

Class imbalance is one obvious candidate explanation for the performance decrease: as the number of examples $N$ in the dataset grows and the number of examples for the new symbol $\hat{y}$ stays fixed, the prior probability of the new symbol $p(\hat{y}) \sim \frac{\text{count}(\hat{y})}{N}$ in the training data decreases. 
If this were true, simply upsampling the new symbol's annotations should revert the decrease. 
With a view to addressing class imbalance, we explore two common solutions: upsampling and group distributionally robust optimization (DRO) \citep{sagawa.s.2019, sagawa.s.2020}. 
However, neither of them fully attenuates the accuracy drop nor delivers a satisfactory solution. 

The failure of these solutions leads us to identify a different force associated with the performance decrease, \textit{source signal dilution}, whereby the reliability of the signal coming from indicative tokens in the user utterances for the new symbol is increasingly diminished in larger datasets.
\begin{figure}
    \centering
    \includegraphics[width=\linewidth]{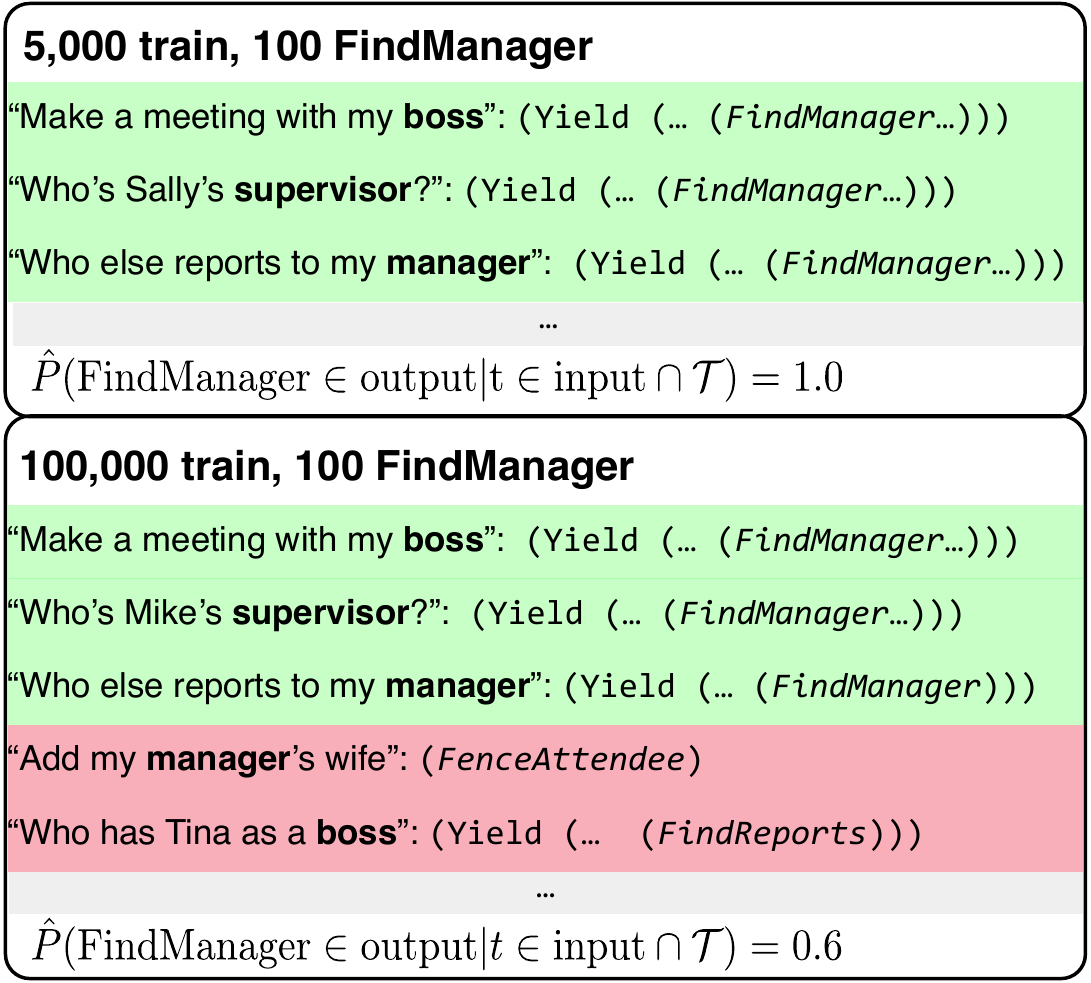}
    \vspace{-2.0em}
    \caption{Source signal dilution in the training set: as the dataset grows, the set of cues $t$ (in bold) associated with \mytt{FindManager} becomes less predictive of it.}
    \label{fig:dilution}
    \vspace{-1.5em}
\end{figure}
This force is illustrated in \cref{fig:dilution}.
At low data settings, some tokens are highly correlated with the \mytt{FindManager} symbol, but as the dataset grows, the correlation with these tokens is diluted by competing examples that, often by coincidence, contain the same tokens. 
We show that when diluting examples (shown in red) are removed, the accuracy drop largely disappears, indicating that our state-of-the-art neural NLU models are overly-reliant on simple lexical cues for learning the new symbol.
Thus, when such lexical cues become less reliable in larger datasets, the models struggle to learn the new symbol.
We later argue that while the removal of diluting examples may address the symptoms expressed under increasing dataset sizes, it falls short of an adequate or scalable solution. 

\noindent Our main contributions are threefold:
\begin{itemize}[noitemsep,nolistsep,topsep=0pt]
    \item We identify a troubling quirk in developing broad-coverage NLU systems that challenges the common assumption that more data typically entails better performance. 
    \item Based on our observations, we identify plausible forces leading to the decreased accuracy seen in \cref{fig:intent_and_calflow}, foremost among them the dilution of the source signal (cf.\ \cref{fig:dilution}). A deeper understanding of this force may guide systematic solutions to this problem. 
    \item Finally, we release our code and models, including a model for the SMCalFlow dataset \citep{andreas.j.2020} that sets a new state of the art on the test set.
    We hope it will serve as a useful baseline for future development on this challenging dataset.  
\end{itemize}
 
\section{Datasets and Models}
\vspace{-0.5em}
\label{sec:data_and_model}

Intent recognition and task-oriented semantic parsing share multiple common features.
They both translate user utterances to symbols in a certain meaning representation and they are both commonly used in production NLU systems.
This makes them ideal testbeds for exploring the dynamics in incremental symbol learning.

\vspace{-0.5em}
\subsection{Intent Recognition} \label{sec:intent_data}
Intent recognition involves classifying utterances into a given set of \textit{intents} \citep{lorenc.p.2021}.
Intents often index into a set of pre-defined templates (e.g., the intent \mytt{play\_music} might index into a template with slots ``song name,'' ``song artist'' etc.) and are central to many digital assistant technologies. 
New intents may be added to the agent incrementally during the development process as needs for new capabilities arise. 
For example, an agent capable of cooking tasks may be extended to other household tasks, requiring it to understand the associated intents. 

We use the \emph{NLU evaluation} data provided by \citet{liu.x.2019}, containing \num{25715} utterances for \num{68} intents across \num{18} scenarios. 
\num{2571} and \num{5144} utterances are reserved for validation and testing, respectively. 
When examining each new intent, we fix the number of examples for the new intent at \num{30}, chosen to represent a low-resource intent, and vary the size of the training set $N \in \{\numlist{750;1500;3000;7500;15000;18000}\}$ where \num{18000} is the \textit{max} in \cref{fig:intent_and_calflow}. We examine \num{5} intents:
\begin{itemize}[noitemsep,nolistsep, topsep=0pt]
    \item \mytt{play\_radio} is primarily triggered when users ask for radio stations to be played.
    \item \mytt{email\_query} is for email-related queries.
    \item \mytt{email\_querycontact} is triggered by questions about contacts in an address book. 
    \item \mytt{general\_quirky} is a catch-all category for trivia-style questions and pleasantries.
    \item \mytt{transport\_traffic} is triggered by traffic-related questions and commands. 
\end{itemize} 
Some of the intents (e.g., {\tt{play\_radio}}) have a set of easily-identifiable trigger tokens (e.g., ``radio'' and ``fm'') while others (e.g., {\tt{general\_quirky}}) have very diverse inputs. 
Example utterances for each intent can be found in Appendix~\ref{append:intent}. 

To model this task, we apply a linear classification layer to the {\tt{[CLS]}} token of BERT base \citep{devlin.j.2019}, finetuning the whole contextualized encoder at training time. 
This model was trained to convergence with the Adam optimizer using a learning rate of \num{1e-5}.

\vspace{-0.5em}
\subsection{Semantic Parsing} 
While providing a good environment for experimentation, the intent recognition task lacks the complexity of a full real-world production environment. 
We therefore seek to expand our experiments and analyses to a production-level task and dataset.  
 \begin{figure}
    \centering
    \includegraphics[width=\linewidth]{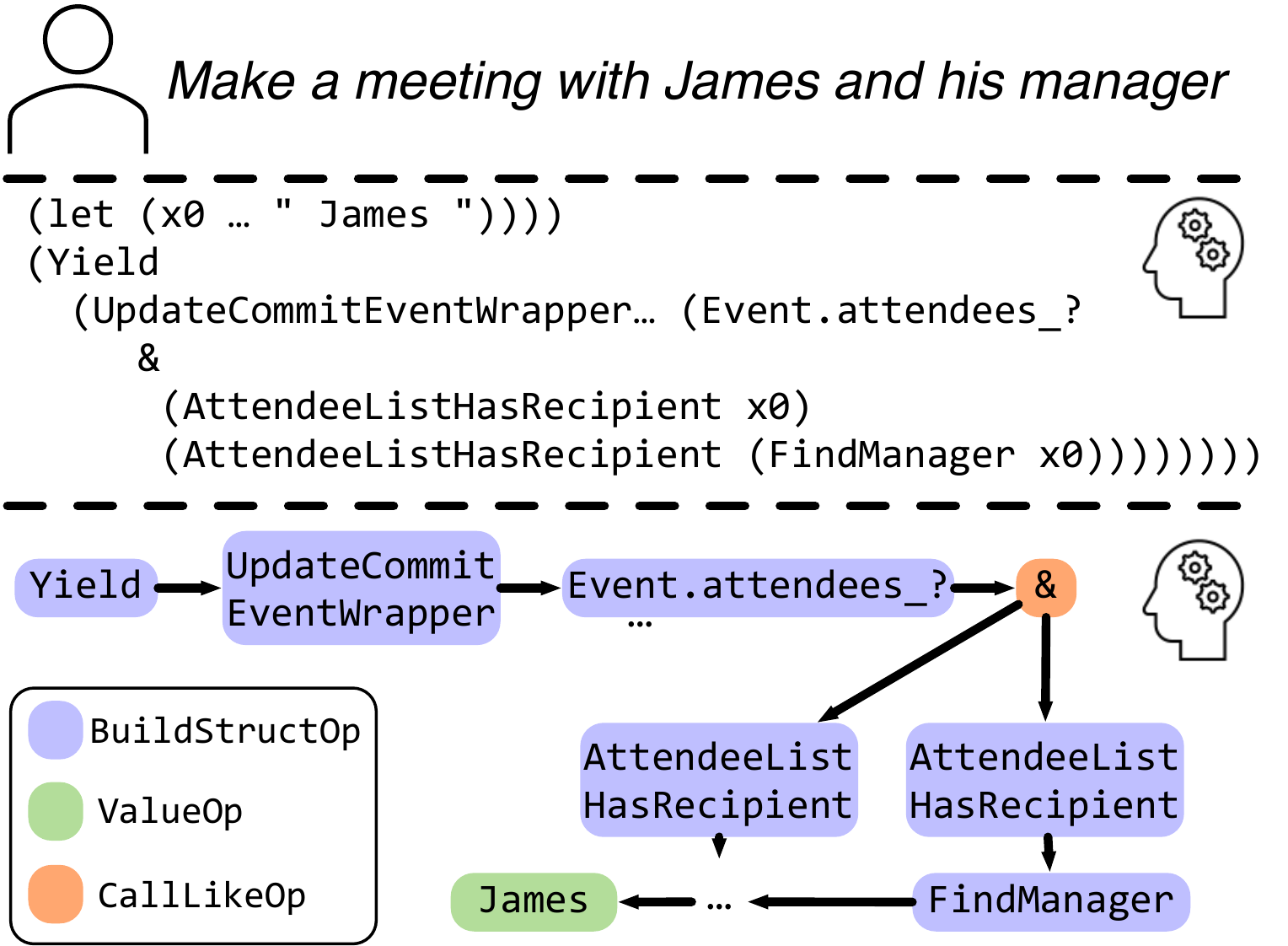}
    \vspace{-1em}
    \caption{Example of an SMCalFlow program; it can be represented as a Lisp expression (middle) or as a DAG (bottom).}
    \vspace{-1em}
    \label{fig:calflow_data}
\end{figure}
To that end, we use the SMCalFlow dataset \citep{andreas.j.2020}, which offers a task-oriented semantic parsing challenge, where a user iteratively creates a dataflow graph in a dialogue with an agent (cf. Fig.~\ref{fig:calflow_data}).
The dataset has \num{41517} dialogues with \num{338} function types, yielding \num{121024} training user turns in the full setting. 
Following prior work~\citep{andreas.j.2020, roy.s.2022}, the input to our parsing model is the previous user utterance, the corresponding agent response, and the current user utterance, all concatenated. 
The model is tasked with learning to generate a typed Lisp program; see Fig.~\ref{fig:calflow_data} for an example, (further examples in Appendix \cref{tab:smcalflow_appendix_examples}).

We explore both a sequence-to-sequence (seq2seq) model and a sequence-to-graph (seq2graph) model, using the MISO framework \citep{zhang.s.2019b, stengel-eskin.e.2021}, which is built on top of AllenNLP \citep{gardner.m.2017}. 
The former directly predicts the Lisp string, while the latter produces a DAG as seen at the bottom of \cref{fig:calflow_data}. Details follow.

\vspace{-0.5em}
\paragraph{LSTM seq2seq} Our baseline model is an LSTM-based seq2seq model similar to that used by \citet{dong.l.2016} and \citet{andreas.j.2020}. 
It consists of a BiLSTM encoder and an LSTM decoder with attention over the encoder states and a source copy operation to copy entity spans from the source text; its embedding layer is initialized with GloVe embeddings \citep{pennington.j.2014}.
See \cref{append:lstm} for more details. 

\vspace{-0.5em}
\paragraph{Transformer-based transductive} A more competitive approach follows the transductive parsing paradigm \citep{zhang.s.2019a}, which aims to directly produce the underlying DAG instead of the surface form, generating graph nodes as well as edges.  
We implement a transformer-based transductive model based on the architecture and code from \citet{stengel-eskin.e.2021}. 
The model directly generates the linearized DAG (cf.\ \cref{fig:calflow_data}) underlying the SMCalFlow Lisp expression; the nodes of the DAG (functions and arguments) are generated in a sequence-to-sequence fashion, with graph edges and edge types being assigned during decoding via a biaffine parser \citep{dozat.t.2016}.  
Following past work, the input features for this model are a concatenation of BERT \citep{devlin.j.2019}, GloVe, and character CNN features. 
We also experiment with RoBERTa \citep{liu.y.2019} and ALBERT \citep{lan.z.2019} as the encoder in \cref{append:roberta}, which show qualitatively similar behavior. Therefore, we report BERT results in the remaining experiments. 
We give additional details on the transductive transformer model in Appendix~\ref{append:transformer}.

\begin{figure*}
    \centering
    \includegraphics[width=\textwidth]{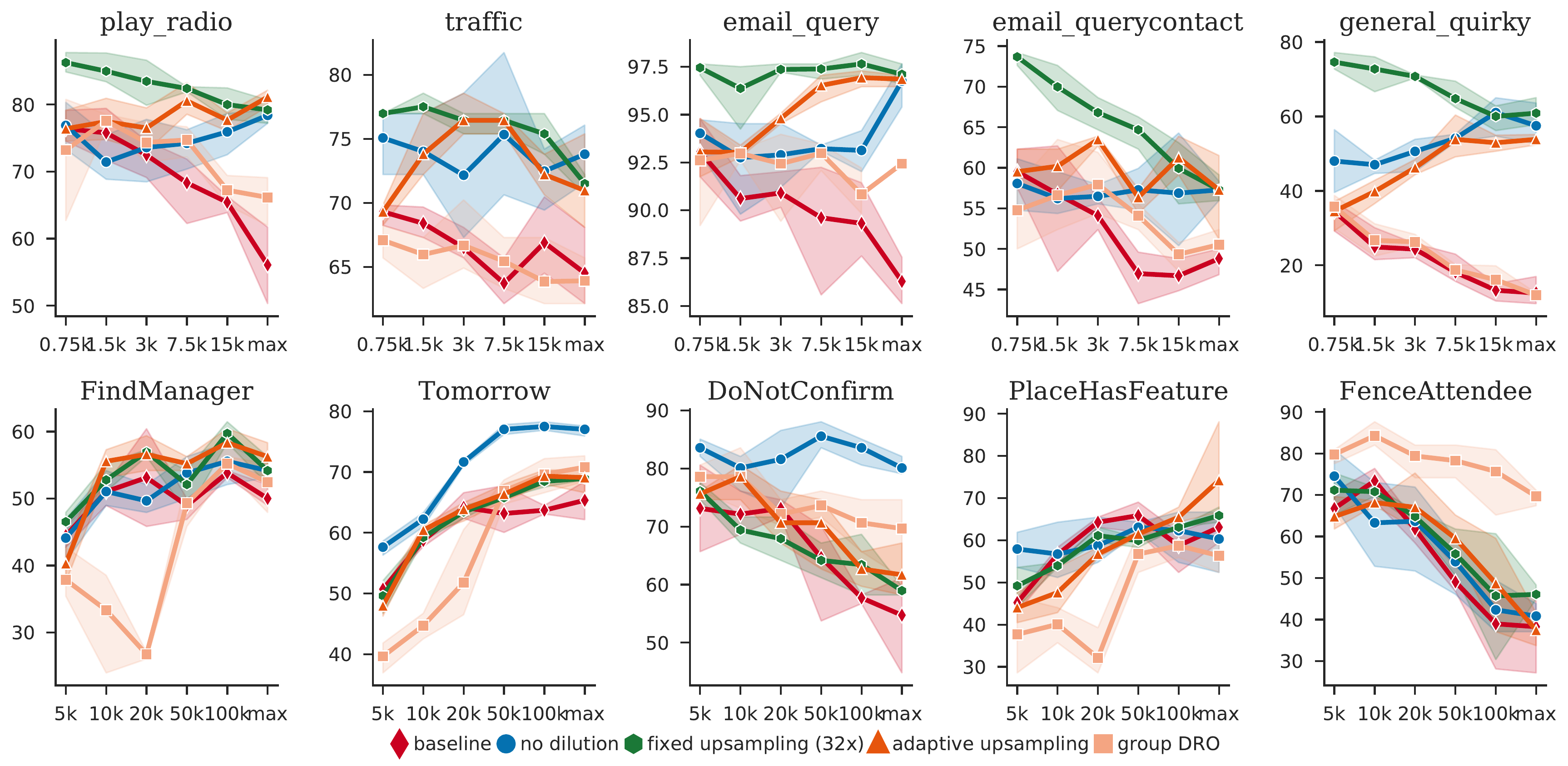}
    \caption{Per-symbol accuracy on intent recognition and semantic parsing as the size of the training set increases. Shaded regions represent 95\% bootstrap confidence intervals.}
    \label{fig:big_fig}
\end{figure*}

\vspace{-0.5em}
\paragraph{Data} As SMCalFlow's test set is not publicly available, we first split the validation data in half to obtain a held-out test set. 
We then construct training splits by selecting 100 examples for each new symbol and varying $N \in \{\numlist{5000;10000;20000;50000;100000}, \text{\emph{max}}\}$, where \emph{max} is the maximum amount of data that can be taken from the $\num{121024}$ training examples while excluding all but $\num{100}$ examples for the new symbol. 
This number is typically slightly under $\num{120000}$.
We describe the symbols examined; Appendix~\ref{append:calflow_data} has further examples.  
\begin{itemize}[nolistsep,noitemsep,topsep=0pt]
    \item \mytt{FindManager} is invoked when a user queries for a person's manager.
    \item \mytt{Tomorrow} returns tomorrow's date. 
    \item \mytt{DoNotConfirm} is applied when a user wants to cancel a proposed action (e.g., confirming the creation of an event) by the agent. 
    \item \mytt{FenceAttendee} is invoked when a user tries to create an event with a person without using their proper name (e.g., ``mom'' or ``my dentist'').
    \item \mytt{PlaceHasFeature} is used when a user asks whether a place has certain amenities (e.g., outdoor seating) or offers certain services. 
\end{itemize}

These vary in compositionality: \mytt{FindManager} and \mytt{PlaceHasFeature} take arguments, but \mytt{Tomorrow} does not, and \mytt{FenceAttendee} and \mytt{DoNotConfirm} are complete programs in themselves. 
\mytt{FindManager} and \mytt{Tomorrow} have strong input cues, but \mytt{DoNotConfirm} and \mytt{FenceAttendee} come from very diverse inputs.

\vspace{-0.5em}
\section{Experiments}
\vspace{-0.5em}
\paragraph{Baseline} The first experimental setting is the baseline setting, as presented in Fig.~\ref{fig:intent_and_calflow}. 
Here, we vary the size of the training data while holding the number of examples for each new symbol fixed. 
We train each model with three random seeds for all experiments, reporting the average.
Note that we train each model on \emph{all} of the training data (for new and old symbols) simultaneously, which is the typical setting for production systems. 

\vspace{-0.5em}
\paragraph{Upsampling} In our upsampling experiments, we increase the number of examples for a new symbol by duplicating training examples for that symbol. 
While this addresses the class imbalance problem, it does not increase the diversity of inputs.
We explore two ways of setting the number of times we duplicate the existing examples: \textit{adaptive} and \textit{fixed}. 
For adaptive upsampling, we upsample the new symbol examples so that the ratio of new symbol examples to total training size (i.e., $\frac{\text{count}(\hat{y})}{N}$ where $\text{count}(\hat{y})$ is the number of examples for the new symbol $\hat{y}$ after upsampling) remains constant as $N$ increases. 
For example, the ratio for intent recognition is \num{30} new symbol examples in \num{750} total examples (\num{0.04}) so at \num{15000} total examples we would have \num{600} new symbol examples, upsampled from the original \num{30}. 
Note that although this strategy gives us a more comprehensive picture as it keeps class imbalance constant, it is unrealistic in practice: whenever we add new training data for a symbol (often, daily in deployed systems) we would have to proportionally increase the ratio for all other symbols, rapidly expanding the training set. 
Thus, we also examine a fixed strategy, where we upsample by the same ratio as $N$ increases.
After exploring ratios in $\{\numlist{2;4;8;16;32;64}\}$, we selected \num{32} based on validation performance; i.e., each example for the new symbol is copied 32 times in the training set. 
Upsampling eliminates (adaptive) or alleviates (fixed) class imbalance, while also improving the reliability of trigger tokens for the new symbol (e.g., ``manager'' for \texttt{FindManager}) to some degree. 

\paragraph{Group DRO} has been proposed as a method for robust generalization under severe class imbalances \citep{sagawa.s.2019}. 
Rather than optimizing by minimizing the average loss across a training batch, group DRO seeks to minimize the loss for the worst-performing group in each batch. 
More formally, given a set of groups $\mathcal{G}$, a parameter space $\Theta$, a model $f(x; \theta)$, a loss $l$, and a per-group training distribution $P_g$, the group DRO objective is:
\vspace{-0.5em}
\begin{align*}
 \theta^{*} &= \argmin\limits_{\theta \in \Theta} \Big(\max\limits_{g \in \mathcal{G}} \mathbb{E}_{(x,y) \sim P_{g}} \big[l(f(x;\theta), y)\big]\Big)
 \vspace{-0.5em}
\end{align*}
We apply this objective to our intent recognition model, treating each intent as a separate group. As long as the worst-performing group is the new intent (e.g., {\tt{play\_radio}}), the model will be optimized solely for that intent. 
For the SMCalFlow setting, applying group DRO is more challenging, as the output is a program containing multiple functions rather than a single class. 
We apply group DRO to SMCalFlow by defining two groups: programs with the new symbol, and those without it. 

\section{Results and Analysis} 
\subsection{Overall Model Performance}
We first evaluate model performance using the original data splits of the two datasets. 
The purpose is to verify that our chosen models are competitive on these datasets and can serve as a solid foundation for subsequent studies on incremental symbol learning.
For intent recognition, the BERT-based classifier trained on the full dataset obtains $90.5\%$ test accuracy, indicating that it is suited to the task.
\cref{tab:parsing} shows the performance of the semantic parsing models on the full SMCalFlow validation and test splits. Here, the test split is the held-out split used in the official SMCalFlow leaderboard. 
To further increase the seq2graph model's performance, we follow \citet{stengel-eskin.e.2020} and unfreeze the top \num{8} layers of the encoder (e.g. BERT, RoBERTa). 
These tuned models outperform \citet{platanios.a.2020}, setting a new state of the art for SMCalFlow. 
To reduce computation we use models with frozen encoders for the following sections. 

\begin{table}[t]
    \centering
    \small
    \vspace{-0.5em}
    \begin{tabular}{@{}lcc@{}}
    \toprule
        \textbf{Model} & \textbf{Dev EM} & \textbf{Test EM} \\
    \midrule
        LSTM (ours) & $66.9\%$ & $52.4\%$ \\
        TFMR +BERT (ours) & $79.3\%$  &  $74.5\%$\\
        \citet{platanios.a.2020} & -- & $75.3\%$ \\
        TFMR +BERT, tuned (ours) & $80.3\%$ & $75.5\%$ \\
        TFMR +RoBERTa, tuned (ours) & $80.8\%$ & $\mathbf{75.7\%}$ \\
    \bottomrule
    \end{tabular}
    \vspace{-0.5em}
    \caption{Exact-match (EM) accuracy on SMCalFlow official test set when trained on the full dataset. The transformer (TFMR) transductive model with encoder tuning sets a new state of the art.}
    \label{tab:parsing}
    \vspace{-1.5em}
\end{table}

\subsection{More Data Can Hurt Performance} 
\paragraph{Intent Recognition} 
Fig.~\ref{fig:intent_and_calflow} and Fig.~\ref{fig:big_fig} show the overall and per-symbol accuracy of a model when the number of examples for a new intent is fixed at \num{30}. 
As the size of the training set increases, the overall accuracy of the model, averaged across all intents, improves. 
However, the accuracy on the new intent decreases. 

\vspace{-0.5em}
\paragraph{Semantic Parsing} 

\begin{figure}[t]
    \centering
    \includegraphics[width=\linewidth]{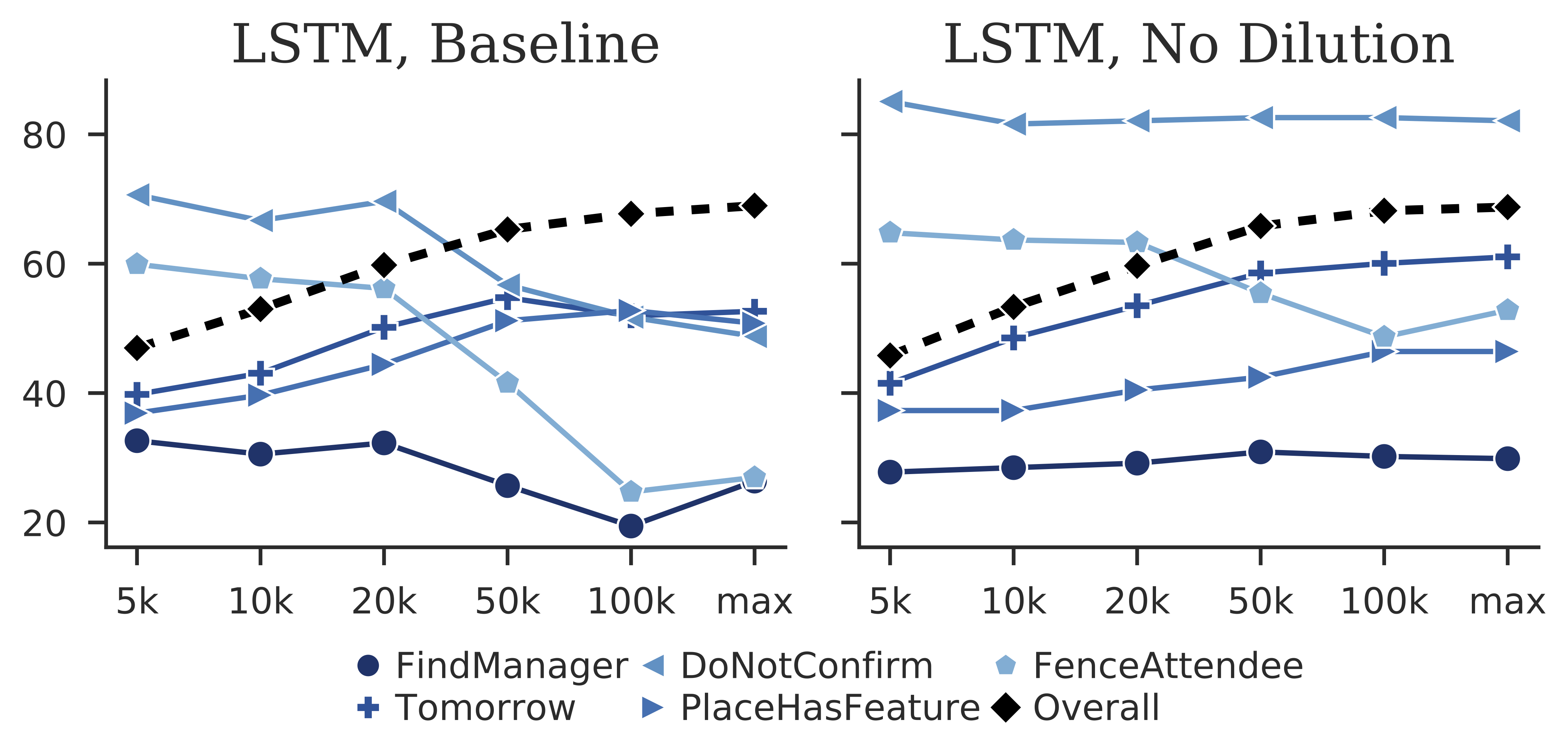}
    \caption{LSTM performance on the new symbol decreases as the total training size increases, but removing source signal dilution largely fixes it.}
    \label{fig:lstm_calflow}
    \vspace{-1em}
\end{figure}

When the number of examples for a symbol is fixed at \num{100}, and the number of other training examples increases, Fig.~\ref{fig:intent_and_calflow} and Fig.~\ref{fig:big_fig} show that the accuracy on new symbols is highly non-monotonic.
Fig.~\ref{fig:lstm_calflow} shows that the non-monotonic accuracy is not just a quirk of transductive parsing but is also seen in a commonly-used LSTM-based seq2seq baseline model.
In \cref{append:roberta} we include results for RoBERTa (base and large) and ALBERT, which indicate that the non-monotonic and often-decreasing SMCalFlow trends are not due solely to the architecture or the encoder. 
While intent recognition displays largely decreasing performance curves, some curves for SMCalFlow symbols (e.g., \mytt{FindManager}) increase and decrease at different settings. 
This may be attributed to competing forces: additional data may increase the seq2seq or seq2graph model's fluency in producing syntactically correct outputs, but it also increases the class imbalance and source signal dilution, with different settings having different balances of these forces.

\subsection{Addressing Class Imbalance} \label{sec:class_imbalance} 
Assuming the number of examples for the new symbol stays fixed, then as the size of the dataset grows, the new symbol represents a smaller and smaller minority as compared to all other symbols combined; i.e., class imbalance increases. 
If the growing class imbalance were the culprit behind the decrease in accuracy observed in Fig.~\ref{fig:intent_and_calflow}, then we would expect a robust optimization technique that prioritizes minority classes, such as group DRO, to ameliorate the problem.
DRO was designed for optimization that is robust to distribution shifts in the group priors between the training and test data. 
In other words, DRO optimizes the model for performance on both majority and minority classes, so that if the minority classes become more prevalent at test time, the overall performance does not suffer. 
In our setting, the test distribution remains fixed, while the training distribution changes as more data is added, with the new symbol becoming less probable, increasing the distribution shift between training and test data.
If the growing class imbalance (i.e., the growing shift between the training and test distributions) were the cause for the decreasing accuracy, DRO would help the model overcome the shift and improve the performance. 

The group DRO curves in Fig.~\ref{fig:big_fig} show that, while the accuracy on the new symbol is often raised above the baseline (e.g., \mytt{FenceAttendee}, \mytt{email\_query}), it generally still decreases as the training set grows. 
Additionally, applying group DRO often results in an accuracy lower than the baseline for many training data sizes (e.g., \mytt{PlaceHasFeature}, \mytt{FindManager}, and \mytt{traffic}). 
Thus, while group DRO may improve the results on the new symbol for a given setting, it does not alleviate the core problem: a larger dataset leads to lower accuracy than what could have been obtained with a smaller dataset. 
This suggests that the underlying problem goes beyond the distribution shift induced by class imbalance. 

The upsampling curves for both strategies (fixed and adaptive) in Fig.~\ref{fig:big_fig} show a similar trend: upsampling can improve overall accuracy and can reduce the rate at which the accuracy decreases, but often fails to remove the decrease. 
In some cases (e.g., \mytt{email\_query}, \mytt{PlaceHasFeature}) fixed upsampling does lead to monotonically-improving accuracy; however, these improvements are inconsistent across symbols, suggesting that there may be other forces at play.

\vspace{-0.5em}
\subsection{Addressing Source Signal Dilution}
The failure of group DRO and upsampling to solve the problem suggests that it may not be due purely to the increased class imbalance. 
We propose another contributing factor: a decrease in the reliability of the source signal when additional data is added. 
For many symbols, there is often a set of tokens $\mathcal{T}$ that can be found in most of the utterances for that symbol. 
For example, for the {\tt{play\_radio}} intent, at least one of the tokens in $\mathcal{T} = \{\texttt{radio}, \texttt{fm}, \texttt{play}\}$ is found in \num{78.04}\% of the corresponding utterances in the full training data. 
Thus, the set $\mathcal{T}$ is a strong signal for predicting {\tt{play\_radio}}. 
However, as more data is added, elements in $\mathcal{T}$ will happen to appear more in the inputs for other intents, reducing their strength as a signal for predicting {\tt{play\_radio}}.
More formally, we define the \textit{source signal strength} of a trigger token set $\mathcal{T}$ to a symbol $\hat{y}$ as $\hat{P} (\hat{y} \in \text{output} \mid \exists t \in \mathcal{T}, t \in \text{input})$, i.e., among the training examples that contain at least one of the trigger tokens in the input, what percentage of those have $\hat{y}$ in the output. \textit{Source signal dilution} then refers to the decrease of the source signal strength of $\mathcal{T}$ to $\hat{y}$ when the overall training data grows, and the trigger tokens become less predictive of the symbol $\hat{y}$.

\begin{figure}[t]
    \centering
    \includegraphics[width=\linewidth]{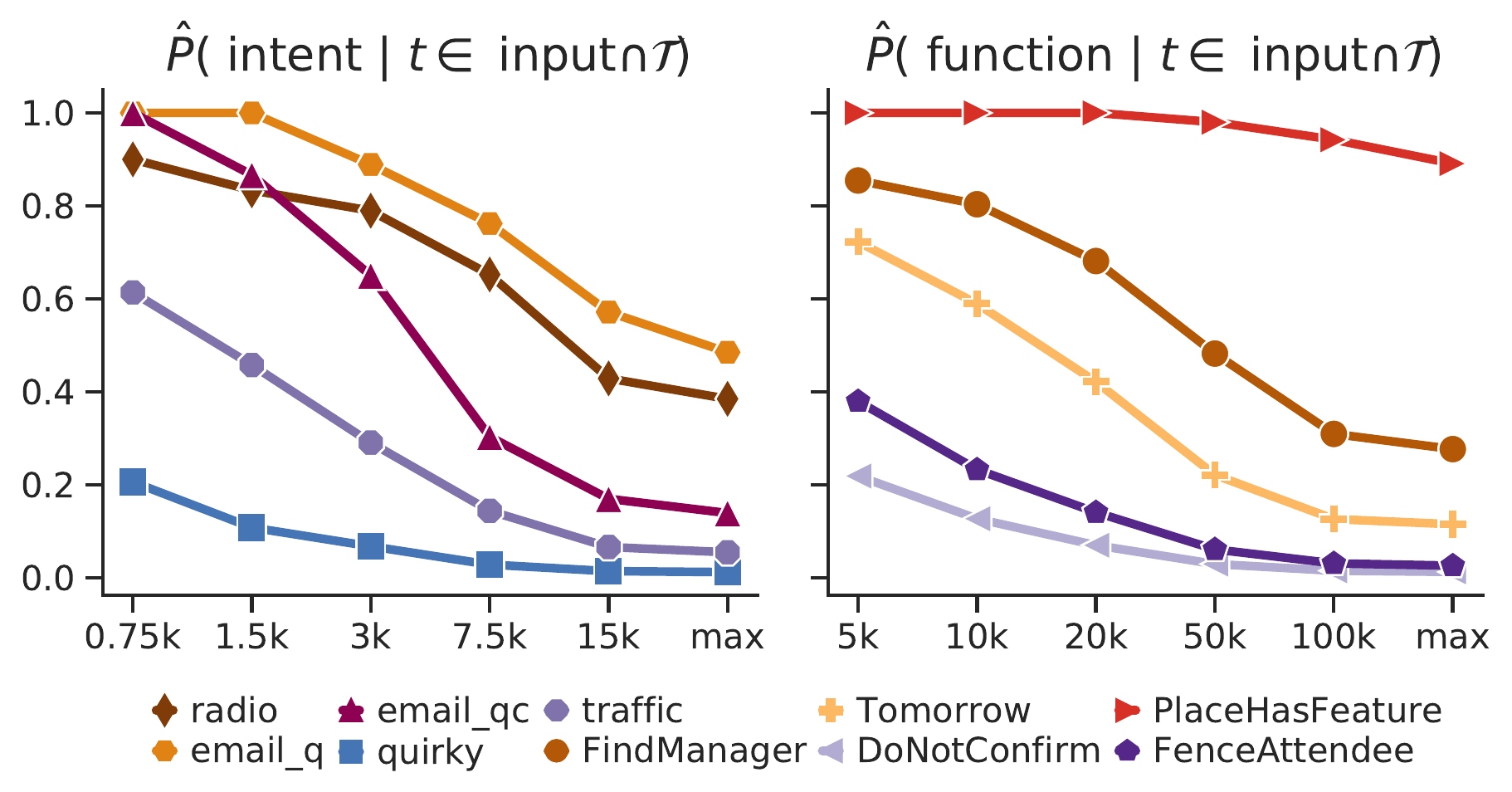}
    \caption{As dataset size increases, the source signal strength of the trigger tokens to each symbol decreases.}
    \label{fig:source_signal}
\end{figure}

\cref{fig:source_signal} shows the source signal strength for each of the examined symbols. 
The set of trigger tokens for each symbol were determined manually and given in Appendix~\ref{append:triggers}. 
Across symbols, as the size of the dataset increases, the source signal strength decreases, with more examples for other symbols containing the same tokens in their inputs, diluting the signal.
Note that the different starting points of these lines indicate that different symbols start with a higher or lower source signal strength.
For example, even at the most concentrated setting (\num{750} total examples to \num{30} \mytt{general\_quirky} examples), \mytt{general\_quirky} has no trigger tokens that are strongly correlated, while \mytt{play\_radio} has a small set of trigger tokens that are perfectly correlated with it.
Taken together with \cref{fig:intent_and_calflow}, \cref{fig:source_signal} shows that source signal dilution has a positive correlation with the decreasing performance on that symbol.  

To further investigate the impact of source signal dilution and see if there is any causal link, we conduct an intervention and experiment with removing the diluting examples (e.g., the ones in red in \cref{fig:dilution}) from the training data. 
In the ``no dilution'' setting of Fig.~\ref{fig:big_fig}, we remove the diluting examples for each new symbol and fill in with other training examples to keep the class imbalance unchanged,\footnote{This holds except for the max setting, where the max amount of data is reduced since more examples are discarded.} but we ensure that the new examples do not contain any of the trigger tokens for the new symbol.
For example, in \cref{fig:dilution}, at $100,000$ training examples, we would not add the two examples highlighted in red.
They would be replaced by examples that do not contain ``manager,'' ``boss,'' and ``supervisor'' in the input.
In other words, we change the datasets such that the curves in \cref{fig:source_signal} remain flat.

Here, we see that the decrease is in fact often attenuated at larger training datasets, and the accuracy even increases for several intents and functions. Considering that the ``no dilution'' setting has exactly the same amount of class imbalance as the baseline setting, this confirms that \textit{source signal dilution is a distinct factor from class imbalance that contributes to the challenges in incremental symbol learning}. 
This result is not unique to BERT-based models: when we remove the same examples for the GloVe-based LSTM SMCalFlow parser, we see similar trends (cf. \cref{fig:lstm_calflow}). 

\vspace{-0.5em}
\subsection{Impact on Competing Examples} 
\begin{figure}
    \centering
    \includegraphics[width=\linewidth]{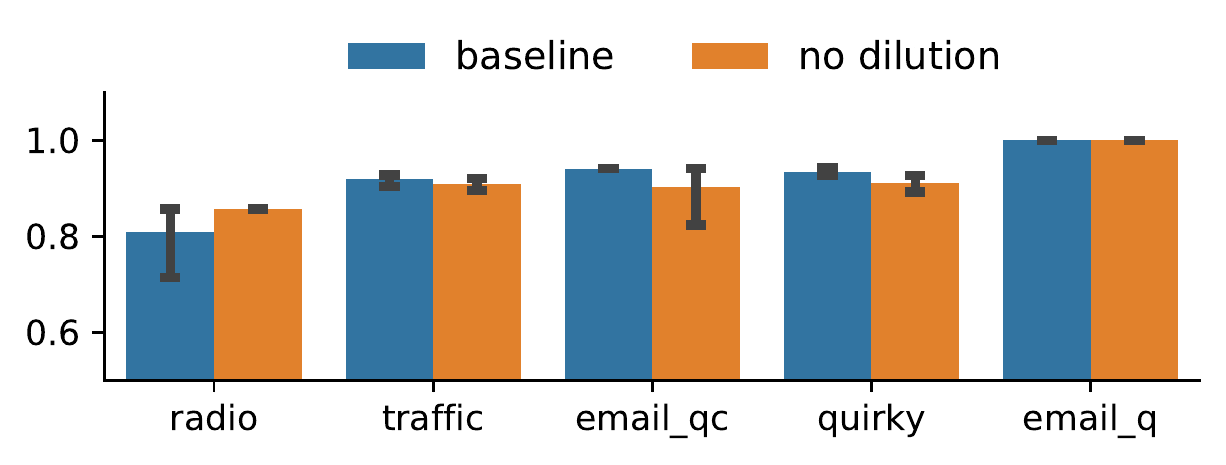}
    \caption{Test accuracy on examples that contain any of the trigger tokens associated with the new intent, but whose label is some other intent.} 
    \label{fig:difficult}
\end{figure}

The increase in performance on the new symbol that results from removing diluting examples, as seen in Fig.~\ref{fig:big_fig}, might come at a cost. 
Specifically, since we are removing examples containing trigger tokens associated with the new symbol (e.g., examples containing ``manager'' in the input but not \mytt{FindManager} in the output), we run the risk of losing performance on those examples in the test set. 
\cref{fig:difficult} shows the intent recognition performance on such test examples (i.e., examples that would have been excluded from the training set), for the maximum number of total training examples (18000).\footnote{SMCalFlow lacks enough examples with source-side competition for other symbols in the test set to perform a similar quantitative evaluation.}
The accuracy decreases when we remove diluting examples, sometimes substantially; i.e., the improvement on the new symbol from removing diluting examples comes at a cost to exactly the type of examples we are removing.
While this is unsurprising, it is unfortunate, as it indicates that the strengthening of the source signal through removal, while perhaps addressing the symptoms of the problem seen in \cref{fig:intent_and_calflow}, falls short of presenting a satisfactory cure.

\section{Discussion} 
Firstly, our results show that, as the training dataset grows, the performance on a new symbol with a fixed number of examples often decreases, suggesting that an increasing amount of data for the new symbol will be required as the dataset expands.
The longer the life-span of a system, the more new symbols may be added to it, meaning that as an NLU system extends its coverage, the data annotation cost may spiral up. 
Simple solutions like upsampling and group DRO do not suffice in this case: even with these in place, the performance remains decreasing or non-monotonic. 
Our results demonstrate that removing diluting examples largely removes the performance decrease. 

However, treating the removal of diluting examples as a solution is unsatisfying in three ways. 
First, we now face a new challenge as we iterate the addition of new symbols. 
Perhaps for the first symbol added, we can successfully remove offending examples from the training set; but as we iterate this process, we may find ourselves removing increasing percentages of our training data, with increasingly disparate subsets of annotations for each symbol. 
This makes removal an unattractive solution. 
Secondly, while we can intervene on the training distribution, we cannot control the user distribution. 
\cref{fig:difficult} shows that the performance on test examples containing trigger tokens but labeled with other intents does decrease after removal. 
This suggests that the model's ability to capture the full range of user utterances may be reduced on some axes after removal, even if the accuracy on the new symbol is increased. 
Finally, treating removal of diluting examples as a solution means accepting that our models rely overly on (and are fooled by) simple lexical cues when learning from a small number of examples. 
Despite the fact that our models leverage the power of recurrence and multi-head attention, and often use large pre-trained contextualized encoders, this reliance on simple non-contextual cues is reminiscent of simpler non-contextual models like Naive Bayes classifiers. 
We would ideally hope that a contextualized representation would be sensitive to the difference between, for example, ``Who is my manager?'' and ``Invite my manager's wife.''
We speculate that the model's difficulty in successfully learning examples such as these simultaneously may be explained a sensitivity to the presence and absence of individual tokens (e.g. ``manager'') rather than a compositional and contextualized analysis of the input;
we leave further examination of this hypothesis to future work. 

Despite these unsatisfying observations, the removal of diluting examples also leads to a more promising conclusion, namely that the models we investigate seem to be able to handle a high degree of class imbalance, provided that the source signal is strong enough for the minority class. 
In the ``no dilution'' setting, the class imbalance remains exactly the same as for the baseline and group DRO settings. 
The strong performance under the ``no dilution'' setting suggests that the models can cope with large class imbalances (e.g., \num{100000} total examples to \num{100} new symbol examples) provided that the lexical cues for the minority class in the training data are strong. 
Both the intent classifier and the transductive parser are capable of handling extremely large class imbalance ratios if the source-target mapping is reliable. 
This helps explain why the models' performance can improve even as the class imbalance increases.

\vspace{-0.25em}
\section{Related Work}
\vspace{-0.25em}
Our learning setting relates closely to work on learning with imbalanced data as well as analyses of spurious correlations.
\citet{sagawa.s.2020} find that over-parameterized networks display a similar trend to our trends on the worst-performing group as model capacity is increased: minority-group performance decreases as overall performance increases. 
They conclude that large models tend to memorize minority-class data and rely on spurious correlations, leading to worsening accuracy.
Our work examines the accuracy on specific symbols as the size of the \emph{dataset} (rather than of the model) grows.
Different solutions have been proposed to improve generalization on minority data, such as distributionally robust optimization \citep[DRO;][]{oren.y.2019, sagawa.s.2019, zhou.c.2021} as well as other training and re-weighting strategies \citep{liu.e.2021, ye.h.2021}. 
These solutions are typically applied to image classification tasks; in a space more closely related to NLU,  \citet{li.j.2014} explore several upsampling and re-weighting strategies for discourse relation classification with imbalanced data, and \citet{larson.s.2019} investigate the effect of imbalanced data for detecting out-of-scope intents. 
\citet{gardner.m.2021} argue that simple lexical features, such as the ones we highlight, represent spurious correlations in the data; the models investigated here would be prone to over-reliance on correlations considered spurious by that account, with the removal of diluting examples strengthening them. 
In a similar vein, \citet{mccoy.t.2019} present evidence that natural language interface models rely upon spuriously-correlated features, and present a challenge dataset with such correlations mitigated.
\citet{yaghoobzadeh.y.2021} connect a model forgetting an example during training (i.e., the example was correctly classified, and then became incorrectly classified) with spurious correlations and propose a two-stage finetuning solution for improving model robustness. 

This past work in learning with spurious correlations and imbalanced data has focused on single-label multi-class classification problems; we follow this trend in our experiments with intent recognition. 
However, we go beyond the single-label setting in our semantic parsing experiments, where we investigate class imbalance in a highly structured multi-label multi-class output space. 

The challenging setting we present differs also from never-ending learning \citep{mitchell.t.2018} and domain adaptation/continued training in that for each iteration of the dataset, a new model is trained, rather than continued training on a single model. 
\citet{li.zhuang.2021} investigate few-shot learning for semantic parsing via continued training, where a trained model is exposed to a small set of annotations for a new predicate. 
While we also attempt to learn from relatively few annotations, we do not adapt learned models, instead simulating the common production setting where models are re-trained on datasets as a whole. 

Previous parsing approaches for SMCalFlow have followed both modeling paradigms used here: \citet{andreas.j.2020} present a seq2seq baseline for SMCalFlow, following previous work in seq2seq semantic parsing \citep{vinyals.o.2015, dong.l.2016, jia.r.2016}. 
\citet{platanios.a.2020} outperform that baseline with a transductive seq2graph model using explicit type constraints. 
Treating semantic parsing as a a seq2graph transduction problem has proved to be a strong paradigm for parsing Abstract Meaning Representations \citep{banarescu.l.2013, zhang.s.2019a, zhang.s.2019b}, Semantic Dependencies \citep{oepen.s.2014, oepen.a.2016, zhang.s.2019b},  Universal Conceptual Cognitive Annotation \citep{abend.o.2013, zhang.s.2019b}, Universal Decompositional Semantics \citep{white.a.2020, stengel-eskin.e.2020, stengel-eskin.e.2021}, and GQA \citep{hudson2019gqa, li.zhuowan.2021}.

\section{Conclusion} 
We examined the effect of a growing dataset on the ease of learning new symbols for NLU, finding that it often becomes harder to learn a new symbol as more data is collected. 
This trend holds across models and settings, and could pose significant problems as NLU systems increase in lifespan and coverage. 
We found that the weakening of simple lexical associations as the datasets grow is closely tied to the decrease in performance, indicating that the neural models tested in this study may be overly reliant on simple lexical cues. 
We end by encouraging others to examine these effects in the problems tested here and also in similar problems, where similar effects are likely to be found. 

\section{Limitations}
In our work, we have examined intent recognition and semantic parsing; while these tasks are prototypical NLU tasks, they represent a subset of natural language processing tasks in which the challenges associated with incremental symbol learning might appear.
Furthermore, we consider two datasets from these tasks, which we deem to be good representatives but which we acknowledge are not exhaustive.
Importantly, as we state in \cref{sec:intro}, all datasets examined here are in English---while we expect the results to hold for other languages, this remains to be verified empirically.
Specifically, our investigation relies heavily on \emph{tokens} to measure source signal dilution; this may need to be modified for languages that have more or less semantic content per token than English.

\section*{Acknowledgements}
We would like to thank the reviewers for their comments and suggestions. 
Elias Stengel-Eskin is supported by an NSF Graduate Research Fellowship.

\bibliographystyle{acl_natbib}
\bibliography{incremental_clean}

\appendix

\begin{appendix}
\section{Data} 
\subsection{Intent recognition} \label{append:intent}
Table~\ref{tab:intent_examples} contains example utterances for each intent. 

\begin{table*}[ht]
    \centering
    \small
    \rowcolors{2}{gray!25}{white}
    \begin{tabular}{p{0.28\linewidth} p{0.72\linewidth}}
    \rowcolor{gray!50}
    \hline
    \textbf{Intent} & \textbf{Utterance} \\
    \hline
    \mytt{play\_radio} &  \emph{play radio mirchi for me} \\
    \mytt{play\_radio} &  \emph{go to channel one hundred and six point nine} \\
    \mytt{play\_radio} &  \emph{i want to hear morning edition on npr}  \\
    \mytt{play\_radio} &  \emph{are you set radio on my favorite radio station} \\
    \hline 
    \mytt{email\_query} &  \emph{open email for unread mails} \\
    \mytt{email\_query} &  \emph{what is the subject of latest email i got and who sent it} \\
    \mytt{email\_query} &  \emph{has dad sent any emails recently} \\
    \mytt{email\_query} &  \emph{new email from mom} \\
    \hline 
    \mytt{email\_querycontact} & \emph{find all the contacts named john} \\
    \mytt{email\_querycontact} & \emph{what is mary s.'s birthday} \\
    \mytt{email\_querycontact} &  \emph{what information do you have on file in my information about bill} \\
    \mytt{email\_querycontact} & \emph{give me charles telephone number} \\
    \hline 
    \mytt{general\_quirky} & \emph{nice to talk to you} \\
    \mytt{general\_quirky} & \emph{ask me an arithmetic question} \\
    \mytt{general\_quirky} & \emph{i would like it to help with coding debugging} \\
    \mytt{general\_quirky} & \emph{i like my robot to talk to me like a friend} \\
    \hline 
    \mytt{transit\_traffic} &  \emph{what is the traffic situation right in broadway street}  \\
    \mytt{transit\_traffic} &  \emph{what is the traffic like today} \\
    \mytt{transit\_traffic} &  \emph{is there traffic right now in maiden lane} \\
    \mytt{transit\_traffic} &  \emph{let me know about current traffic in carmen drive} \\
    \bottomrule
    \end{tabular}
    \caption{Examples of intent recognition data.}
    \label{tab:intent_examples}
\end{table*}
\subsection{Semantic parsing} \label{append:calflow_data}
The SMCalFlow data consists of user-agent dialogues, where the agent produces executable Lisp programs based on user commands. 
Variable binding can be performed in Lisp to refer to a value multiple times in a program in a parsimonious way. 
Underlyingly, the variable binding procedure corresponds to re-entrancy in the DAG encoding the program graph. 
\begin{table*}[ht]
    \centering
    \begin{tabular}{@{}p{0.25\linewidth} p{0.75\linewidth}@{}}
    \hline
    {\bf Symbol} & {\bf Tokens} \\
    \hline
     \mytt{email\_query} & emails, inbox \\
     \mytt{email\_querycontact} & contact, phone, number \\
     \mytt{general\_quirky} &  day, today, tell, can\\
     \mytt{play\_radio} &  channel, radio, fm, point, station, tune \\
     \mytt{transport\_traffic} &  traffic \\
     \hline
    \mytt{FindManager} & boss, manager, supervisor \\ 
    \mytt{PlaceHasFeature} & takeout, casual, waiter \\
    \mytt{Tomorrow} & tomorrow \\
    \mytt{FenceAttendee} & meet, mom \\
    \mytt{DoNotConfirm} & cancel, n't, no \\
    \hline
    \end{tabular}
    \caption{Triggers for each symbol. Triggers were chosen manually by inspecting the list of tokens that yielded high $\hat{P}(\text{symbol} | t \in \text{input} \cap \mathcal{T})$ the lowest data setting for each function.}
    \label{tab:triggers}
\end{table*}
Thus, the SMCalFlow parsing task can be tackled either at the level of the Lisp string (sequence-to-sequence) or at the level of the DAG (sequence-to-graph), with the latter approach demanding a method for handling re-entrant nodes in a graph. 
\cref{tab:smcalflow_appendix_examples} shows examples for each SMCalFlow function considered. 

\begin{table*}[h]
    \centering
    \small
    \rowcolors{2}{gray!25}{white}
    \begin{tabular}{p{0.15\linewidth} | p{0.35\linewidth} | p{0.35 \linewidth}}
    \rowcolor{gray!50}
    \hline
    \textbf{Function} & \textbf{Dialogue Context} & \textbf{Current User Utterance}  \\
    \hline
    \mytt{FindManager} & N/A & \emph{Make an event with Abby and her boss} \\
    \mytt{FindManager} & \textbf{User}: \emph{Who are Jake's reports}, \textbf{Agent}:  \emph{Jake Cobb has no direct reports.} & \emph{Who does he report to?} \\
    \mytt{FindManager} & \textbf{User}: \emph{Add an event called presentation with Jamal and his supervisor for Friday at 11.} \textbf{Agent}: \emph{Is this good?} & \emph{Add Igor and his supervisor to this as well.} \\
    \hline 
    \mytt{Tomorrow} & N/A & \emph{Find an event for tomorrow after 4 pm.}\\
    \mytt{Tomorrow} & N/A & \emph{Schedule lunch with Nick tomorrow at noon} \\
    \mytt{Tomorrow} & \textbf{User}: \emph{What time will the sun rise in seattle tomorrow} \textbf{Agent}: \emph{Sunrise will be at 12 : 00 AM tomorrow.} & \emph{what time will the sun set in seattle tomorrow}\\
    \hline 
    \mytt{DoNotConfirm} & \textbf{User}: \emph{Can you change the time to 4 instead?} \textbf{Agent}: \emph{How about now?} & \emph{No, I don't like either of those.} \\
    \mytt{DoNotConfirm} & \textbf{User}: \textit{No I need it to be in the afternoon}, \textbf{Agent} \textit{Does one of these work}? & \emph{No they don't} \\
    \mytt{DoNotConfirm} & \textbf{User}: \emph{Schedule a dentist appointment tomorrow afternoon}, \textbf{Agent}: \emph{Does one of these work?} &  \emph{No} \\
    \hline
    \mytt{FenceAttendee} & N/A &  \emph{Create lunch with mom on sunday} \\
    \mytt{FenceAttendee} &  \textbf{User}: \textit{What events do I have tomorrow}, \textbf{Agent}: \textit{I found 2 events tomorrow}. & \emph{Add my sister, brother, and Daniel} \\
    \mytt{FenceAttendee} & N/A & \emph{Can you tell me if I meet with our repair rep this week or next week?} \\
    \hline 
    \mytt{PlaceHasFeature} & \textbf{User}: \textit{What cuisine do they serve?} \textbf{Agent}: \textit{Sorry, I can't handle that yet}. & \emph{Does the Black Bottle restaurant have a full service bar?} \\
    \mytt{PlaceHasFeature} & \textbf{User}: \textit{Find me Round Table Pizza in Truckee}, \textbf{Agent}: \textit{I found one option}. & \emph{Could I bring a party of people there?}\\
    \mytt{PlaceHasFeature} & N/A & \emph{Is Bamonte's in Brooklyn capable for large parties?}\\
    \bottomrule
    \end{tabular}
    \caption{Example data for SMCalflow semantic parsing.}
    \label{tab:smcalflow_appendix_examples}
\end{table*}

\subsection{Trigger Tokens} \label{append:triggers}
\cref{tab:triggers} has the trigger tokens per symbol. 
These were determined manually by examining tokens which yielded high
 $\hat{P} (\hat{y} \in \text{output} \mid \exists t \in \mathcal{T}, t \in \text{input})$ at the lowest data setting.

\section{Models}
\subsection{LSTM} \label{append:lstm}
The LSTM model takes as input the previous user utterance, the produced agent utterance (if these are available) and the current user utterance, all separated by special tokens.
These are tokenized and embedded using an embedding layer initialized with 300-dimensional GloVe embeddings \citep{pennington.j.2014}. 
Note that there is no contextualized encoder used here. 
The encoder is a 2 layer stacked BiLSTM, with a hidden size of 192 and dropout of $p=0.5$ between cells.
The decoder embeddings are initialized randomly and are also 300-dimensional. 
The decoder also has 2 layers with a hidden size of 384, and recurrent dropout of $p=0.5$. 
The source attention is implemented as an MLP with hidden size 64. 
Batches are bucketed by length during training, and a patience threshold of 20 epochs without improvement is set. 
The LSTM models are trained with ADAM using a learn rate of $1e-3$ and weight decay of $3e-9$. 
Note that for SMCalFlow this paradigm is fairly weak due to its tendency to produce malformed Lisp expressions at lower data regimes and the handling of variable binding through {\tt{let}} expressions.

\subsection{Transformer} \label{append:transformer}
For the transductive model, the DAG for a program (cf. \cref{fig:calflow_data}) is first transformed into a tree by copying and co-indexing re-entrant nodes. 
The tree is then linearized into a sequence of nodes, edge heads, edge types, and node indices. 
At test time, the model produces these sequences, which can be deterministically reconstructed into a DAG by merging co-indexed nodes. 
The generation component of the model maintains a dynamic output vocabulary over three operations: generation from a fixed vocabulary, source copying from the input, and target copying from previously generated tokens. 
The target copy operation allows the model to handle re-entrant nodes, which appear more than once in the linearized tree. This operation allows us to later recover node indices and thus re-build a DAG by merging copied nodes. 
The edge heads and labels are parsed by a biaffine parser \citep{dozat.t.2016}. 
This allows the model to handle functions, arguments, and types separately via typed edges.
Each operation type (\mytt{ValueOp}, \mytt{BuildStructOp}, \mytt{CallLikeOp}) corresponds to a different edge type; the edge types for arguments are also indexed to allow for explicit argument ordering (e.g. \mytt{arg0}, \mytt{arg1}, etc.). 

The input to the model is the same as for the LSTM: the concatenation of the previous two dialogue turns, followed by the current user utterance. These are tokenized and embedded with 300-D GloVe embeddings as well as 100-D character CNN features. There is embedding dropout with $p=0.33$ to prevent overfitting.
The input text is also passed through \mytt{bert-base-cased}, with each subword receiving a 768-D representation. 
These are max-pooled across subword tokens to align with the token-level embeddings. 
The encoder hidden size is 512, with a 8 heads and a feed-forward dimension of 2048. 
The layer-norm and feedforward layers are swapped, and the weight initialization is downscaled by a factor of 512, following \citet{nguyen.t.2019}. 
The encoder has dropout $p=0.2$.

For the transformer, the decoder embeddings are also initialized with GloVe and character CNN features. 
The decoder also has 8 layers with the same dimensions and dropout as the encoder. 
As in the LSTM model, source attention is implemented as an MLP, here with a hidden dimension of 512. The target attention (for target-side copy) is identical. Source attention uses coverage \citep{see.a.2017}. 
The biaffine parser projects the transformer representations to 512 and has dropout $p=0.2$. 
We train transformer models with a patience of 20, using Adam with a linear learning rate warmup stage, followed by exponential learning rate decay. We set the number of warmup steps to 8000. 

\section{Other Encoders} \label{append:roberta}
\begin{figure}
    \centering
    \includegraphics[width=\linewidth]{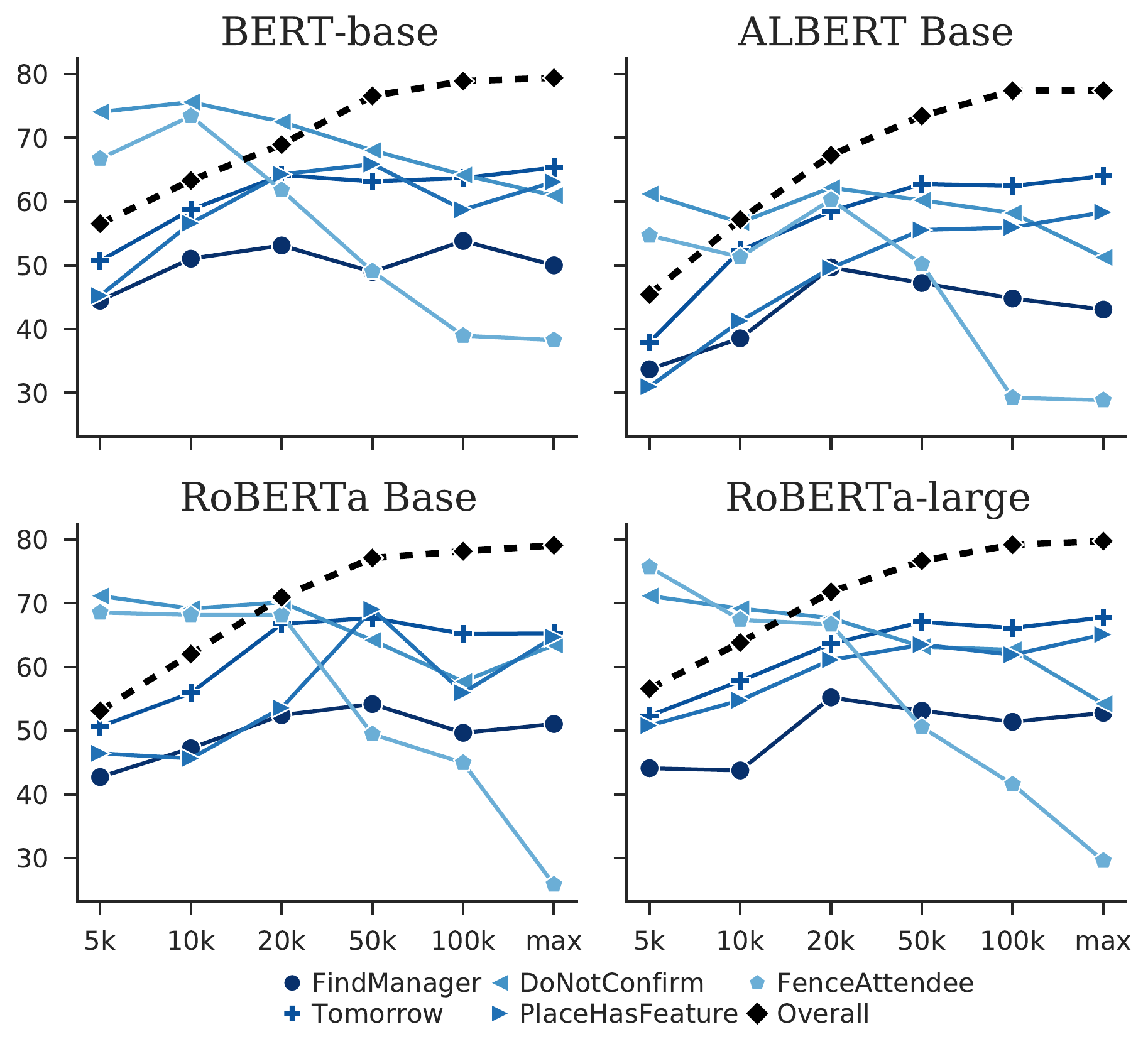}
    \vspace{-1.5em}
    \caption{Parsers with RoBERTa and ALBERT features show similar trends as with BERT features.}
    \label{fig:roberta_albert}
    \vspace{-1.5em}
\end{figure}

\cref{fig:roberta_albert} shows that the trends seen in \cref{fig:intent_and_calflow} also apply also to other encoders, namely ALBERT and RoBERTa (large and base). 
Here, we see that the trends almost exactly mimick the trends seen when using BERT in \cref{fig:intent_and_calflow}. 

The decreasing and often non-monotonic performance as well as the strong similarity of the trends suggests that the source signal dilution problem is not unique only to BERT or to LSTMs.

\section{Licensing} All code is released under an MIT license.

\section{Artifacts} All pre-existing artifacts were used in accordance with their original purpose. The artifacts produced by this paper (code, models) are intended to be used for intent recognition and semantic parsing.

\end{appendix}

\end{document}